\documentclass{article}

    \PassOptionsToPackage{numbers, compress}{natbib}
\usepackage[final]{neurips_2024}

\usepackage[utf8]{inputenc} % allow utf-8 input
\usepackage[T1]{fontenc}    % use 8-bit T1 fonts
\usepackage[backref]{hyperref}       % hyperlinks
\usepackage{url}            % simple URL typesetting
\usepackage{booktabs}       % professional-quality tables
\usepackage{amsfonts}       % blackboard math symbols
\usepackage{nicefrac}       % compact symbols for 1/2, etc.
\usepackage{microtype}      % microtypography
\usepackage{xcolor}         % colors
\usepackage{graphicx}
\usepackage{cleveref}

\title{Adversarial Robustness Analysis of Vision-Language Models in Medical Image Segmentation}

\author{%
  Anjila Budathoki, Manish Dhakal \\
  Department of Computer Science\\
  Georgia State University\\
  Atlanta, GA \\
}

\begin{document}

\maketitle

\begin{abstract}
  Adversarial attacks have been fairly explored for computer vision and vision-language models. However, the avenue of adversarial attack for the vision language segmentation models (VLSMs) is still under-explored, especially for medical image analysis. 
    Thus, we have investigated the robustness of VLSMs against adversarial attacks for 2D medical images with different modalities with radiology, photography, and endoscopy. The main idea of this project was to assess the robustness of the fine-tuned VLSMs specially in the medical domain setting to address the high risk scenario. 
    First, we have fine-tuned pre-trained VLSMs for medical image segmentation with adapters. 
    Then, we have employed adversarial attacks---projected gradient descent (PGD) and fast gradient sign method (FGSM)---on that fine-tuned model to determine its robustness against adversaries.
    We have reported models' performance decline to analyze the adversaries' impact.
    The results exhibit significant drops in the DSC and IoU scores after the introduction of these adversaries. Furthermore, we also explored universal perturbation but were not able to find for the medical images.
    \footnote{https://github.com/anjilab/secure-private-ai}
\end{abstract}

\section{Introduction}
\label{sec:intro}

% Introduction:

% Restate the project topic and objectives.
% Provide background and explain the relevance of your work in the context of AI security and privacy.

% \textcolor{red}{Why adversarial attacks in medical image analysis??}

Artificial Intelligence (AI), especially deep neural networks, is rapidly becoming a pervasive and integral part of everyday applications, including conversational interfaces, decision support systems, and key sectors like education, healthcare, and finance~\cite{chiang2024enhancing, de2023optimized, xiao2023evaluating}. Among these, healthcare stands out as a domain that extensively benefits from AI, spanning applications such as disease diagnosis, monitoring of various health conditions, genetic analysis, and medical image interpretation \cite{thapaliya2024cross, tengnah2019predictive, battineni2020applications, thapaliya2024graph, tang2024prediction, adhikari2023synthetic}. In particular, medical image analysis has seen significant advancements due to deep learning, which has enabled the development of effective assistive diagnostic tools~\cite{wang2025deep, thapaliya2021environmental, kar2024automated}. Given the high stakes of medical decision-making, it is essential that these models demonstrate robustness and reliability—especially since a single false negative in diagnosis can have fatal consequences~\cite{sorin2023adversarial}.

% Aritificial Intelligience, specially Deep neural networks are becoming increasingly ubiquitous, seamlessly integrated into everyday applications such as conversational interfaces, decision support systems, and critical domains like education, healthcare, and finance \cite{chiang2024enhancing, hager2024evaluation, li2022pre, memarian2023chatgpt, lo2023impact, li2023large, de2023optimized, xiao2023evaluating}. Under critical domain, medical image analysis has advanced with assistive tools from the artificial intelligence (AI) community with refinements in deep learning models \cite{wang2025deep, thapaliya2024graph, kar2024automated}.
% Considering the criticality of medical image analysis, the models must be resilient to misclassification; even a single false negative result could be fatal~\cite{sorin2023adversarial}.

Adversarial attacks apply hardly perceptible data perturbation to exploit the blind spots of the trained models, causing the models to maximize the perdiction error~\cite{szegedy2014intriguing}.
These perturbations are not random noise, but calculated modifications to mislead the models.
% These perturbations are typically imperceptible to humans, making defense against them crucial.
The attacks have been studied within the domain of vision-language models (VLMs)~\cite{dong2023adversarial}.
However, there are none to test the robustness of the vision-language segmentation models (VLSMs).
VLSMs are trained to achieve the correct segmentation from images with guidance via text prompts~\cite{luddecke2022image,wang2022cris}.

Considering the criticality of the medical image analysis and the fooling capability of adversarial attacks, there is a need to make the medical VLSMs robust against
the adversarial attacks.
In this research work, as a first step to fill this gap, we study the effects of adversarial images on trained VLSMs.
To validate this effect, we experiment with the multiple modalities of images: endoscopic, radiographic, and photographic images.
The method of this paper can be broken into two major stages: (\textbf{1}) training a VLSM to segment target anatomical regions~\cite{dhakal2024vlsm,poudel2023exploring} and (2) introducing adversarial attacks (PGD~\cite{madry2017towards} and FGSM~\cite{goodfellow2014explaining}).
The comparitive analysis of metrics in the presence and absence of adversarial noise exhibit the vulnerability of the VLSMs models.

\subsection{Vision-Language Segmentation Models}
Vision Language Segmentation Models (VLSMs ) are the models trained to segment an image with guidance from the text prompts~\cite{luddecke2022image,wang2022cris,liu2023gres,Zhou_2023_CVPR}.
The general approach in VLSMs training is to use two different branches of encoders to represent the text and image inputs, and the representations are passed to segmentation mask decoder.
For our project, we have picked CLIPSeg~\cite{luddecke2022image} that has a trained decoder to segment target masks. CLIPSeg uses features from transformers-based image and text encoders of CLIP~\cite{radford2021learning}.

\subsection{Adversarial Attacks}
Adversarial attack introduce the calculated perturbation to the original image such that the model can be fooled to misguide the predictions.
The impact of adversarial attacks in high-stakes domains must be addressed, as they can have serious consequences.  
\cite{finlayson2018adversarial} conducts an experiment across three clinical domains and were successful in both white-box and black box attacks.  Previous research \cite{ma2021understanding} indicates that deep neural network (DNN) models for medical images exhibit greater vulnerability compared to those for natural, non-medical images. 
Adversarial attacks have been extensively studied in the context of vision-language models (VLMs)~\cite{dong2023adversarial, shayegani2023jailbreak, zhao2024evaluating}. 
In this study, we will further investigate the specific task of medical image segmentation.

\begin{figure*}[t]
    \centering
    \includegraphics[width=\textwidth]{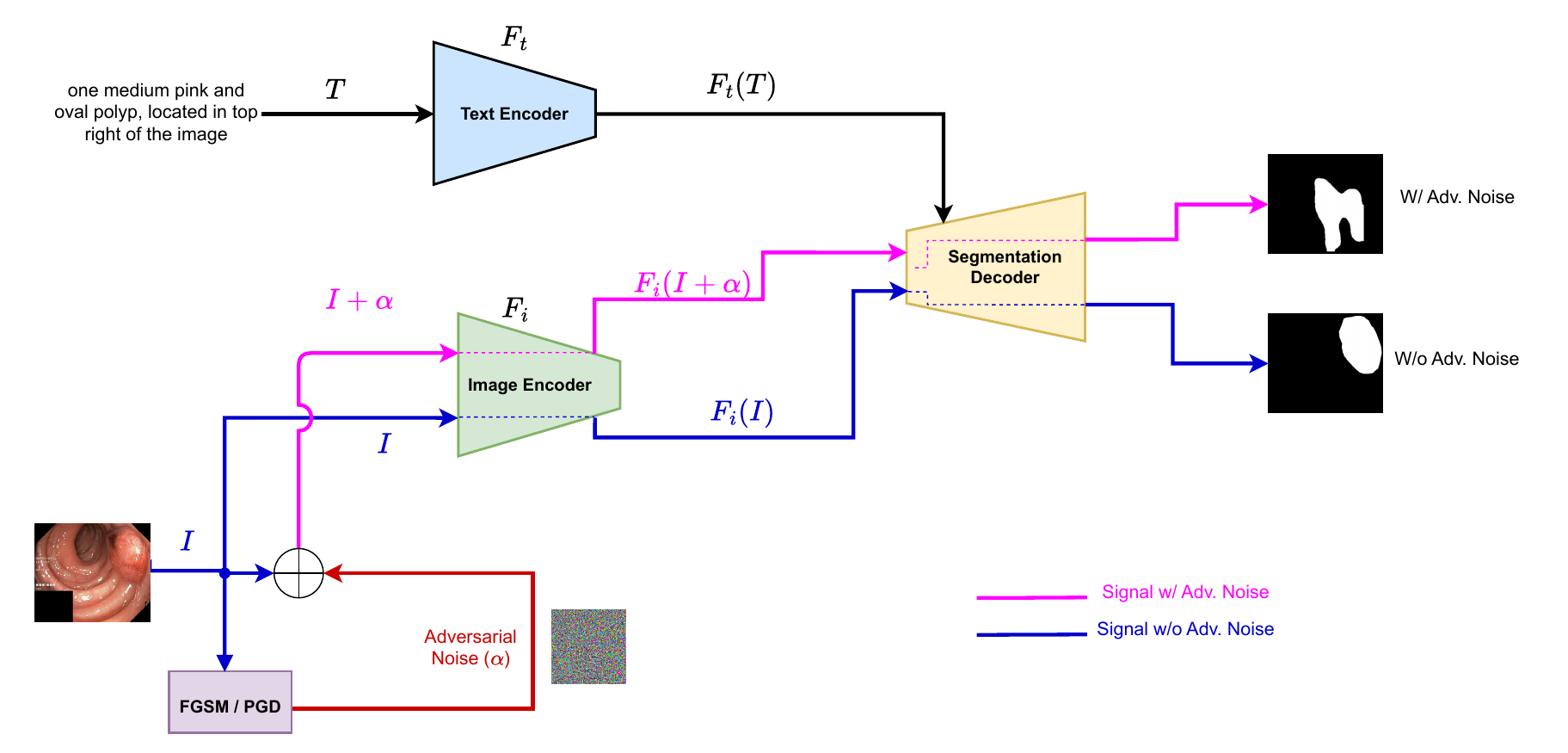}
    \caption{The overall methods of adversarial attack. $F_i$ and $F_t$ are fine-tuned image and text encoders, respectively. By fusing input images with the adversarial noise $\alpha$ generated from FGSM/PGD methods, we observe more inaccurate segmentation mask.}
    \label{fig:proposed_method}
\end{figure*}

\section{Methods and Implementation}
\label{sec:method}

An encoder-decoder pretrained model for vision-language segmentation task is fine-tuned with a smaller training set comprising of the triplets: $D=\{(v_i,l_i,m_i)\}_{i=1}^S$.
Here, $S$ is the number of training samples, $v_i$, $l_i$, and $m_i$ represent the image input, text prompt, and target mask of the $i^{th}$ data point, respectively. 
The input images are RGB images and targets are their corresponding binary masks, i.e., $v_i \in \mathbb{R}^{H \times W \times 3}$, and $m_i \in \{0,1\}^{H \times W}$, respectively.

The overall methodology of this research work can be divided into two major stages: fine-tuning pre-trained VLSMs for medical anatomy segmentation (\cref{sec:vlsm_ft}) and introduction of PGD and FGSM as adversarial attacks on those models \cref{sec:adv_attacks}.

\subsection{VLSM Fine-tuning}
\label{sec:vlsm_ft}

CLIPSeg~\cite{luddecke2022image} provides a pretrained end-to-end VLSMs.
This model was trained to segment commonly seen objects—such as \textit{dog}, \textit{house}, \textit{cup}, etc.—from the natural images in the real world.
Without any modification to the pretrained model, it gives inferior performance when tested directly with task-specific applications like medical image segmentation.
Thus, it needs to be fine-tuned. 
Since the encoders of CLIPSeg are large in size, fine-tuning the entire model is expensive.
So, we resort to fine-tuning lightweight adapters~\cite{houlsby2019parameter} embedded within the encoders

Following the convention and training mechanism provided by VLSM-Adapter~\cite{dhakal2024vlsm}, we fine-tune pretrained CLIPSeg model with the addition of adapter blocks~\cite{houlsby2019parameter}.
Even though~\cite{dhakal2024vlsm} has provided different positioning of adapters in the image and text encoders, we use their optimal variant, \textit{VL-Adapter}.
In this variant, adapters are introduced to both image and text encoders shown in the \cref{fig:proposed_method}.

\subsection{Adversarial Attacks}
\label{sec:adv_attacks}
Our assumption for the threat model is we have access to the gradients i.e white-box attack. For both of the methods of attacks, we will perturb the input image modality as in \cref{fig:proposed_method}.
Small changes in input can significantly fool state-of-the-art networks. 
In their work, \cite{madry2017towards} explored the impact of network capacity on adversarial attacks. 
Over time, models have grown to billions of parameters, and in this study, we investigate how fine-tuned, pre-trained vision-language models (VLSMs) perform against two common adversarial attack methods:  
    \subsubsection{Fast Gradient Sign Method (FGSM)}
    FGSM~\cite{goodfellow2014explaining} is one of the techniques for generating adversarial examples that are $L_\infty$ bounded.  
    To generate adversarial examples with FGSM, we compute the gradient of the loss function with respect to the input $x$. 
    This gradient shows the sensitivity of the model's loss towards changes in the input. 
    Signs of these gradients represent the direction of perturbation that ensures maximum inrcease in model's loss. 
    $\epsilon$ scales the perturbation to control the attack. 
    The steps can be formulated as:
    \begin{equation}
        \label{eq:fgsm}
        x_* = x + \epsilon \cdot sign(\nabla_x \mathcal{L} (\theta, x,y)),
    \end{equation}
   Here, $x_*$ is the perturbed input, $\nabla_x \mathcal{L} (\theta, x,y)$ is the gradient of loss with respect to inputs, $\epsilon$ is the small constant that controls the magnitude of the perturbation and $sign(\cdot)$ is the function that gives signs of the tensor. 
    
    \subsubsection{ Projected Gradient Descent (PGD)}
    PGD~\cite{madry2017towards} is an iterative extension of FGSM.
    In the initial iteration, the adversarial input $x_*^{(0)}$ is given as $x$ or a slight noisy $x$.
    At each iteration, the perturbed input gets updated as:
    \begin{equation}
        \label{eq:pgd}
        x_*^{(t+1)} = clamp(x_*^{(t)} + \alpha \cdot sign(\nabla_{x_*^{(t)}} \mathcal{L} (\theta, x_*^{(t)},y), x - \epsilon, x + \epsilon),
    \end{equation}

    where $\alpha$ is the scaling factor of perturbation, $clamp(a,b,c)$ clamps $a$ with the range of $[b,c]$ input within $[x-\epsilon, x+\epsilon]$, and all of the remaining symbols have similar meaning as in \cref{eq:fgsm}.

    The iteration is run for $T$ steps and the final refined perturbed input is $x_* = x*^{(T)}$.
    $T$ is a hyperparameter, which is $40$ in our experiment.
    % \begin{equation}
    %     \label{eq:proj}
    %     Proj_{x,\epsilon}(z) = x + e \frac{}{}
    % \end{equation}
    
    % \begin{equation}
    %     min_\theta \rho(\theta)
    % \end{equation}
    % where, $\rho(\theta) = E_{(x,y) \sim D}[max_ {\delta \in S} \mathcal{L} (\theta, x + \delta, y)].$
    % The inner maximization part $[max_ {\delta \in S}  \mathcal{L} (\theta, x + \delta, y)]$ deals with generating adversarial example and the outer minimization part $min_\theta \rho(\theta)$ is making the neural network robust to these adversarial example. Here, $\theta$ represents network parameters and $\delta$ is the perturbation constrained within set S. 

\subsection{Experimentations}
\subsection{Implementation Setup}
The training and inference of the VLSM and adversarial attack methods are executed in an NVIDIA RTX 4090. 
We use floating-point-16 mixed-precision training with a batch size of 32.
The models are optimized with AdamW~\cite{loshchilov2017decoupled} with a weight decay of $1e-3$.
The learning rate has a linear function to warmup for the first 20 epochs to reach $1e-3$; after 20 epochs, the learning rate decays with a cosine decay function for the next 180 epochs to reach $1e-5$.
We combined dice and binary cross-entropy losses for the objective function, as shown by:

\begin{equation}
    \mathcal{L} = \lambda_d \cdot \mathcal{L}_{Dice} + \lambda_{ce} \cdot \mathcal{L}_{CE},
\end{equation}
where $\lambda_d$ and $\lambda_{ce}$ are hyperparameters; we chose their values for our experiments as $\lambda_d = 1.5$ and $\lambda_{ce} = 1$.

The bottleneck layer's dimension of the adapter is 64.
Adapters are added to both image and text encoder branches of the model.
The image has been resized to $352 \times 352$ for batch processing, and the context size for text input is $77$.

During the noise injection, different scales $\epsilon  \in \{0.01, 0.03, 0.1, 0.5\}$ are used to determine the amount of perturbation to be introduced in the input images.

\subsection{Datasets}
Poudel et al. \cite{poudel2023exploring} published a variety of language prompts grounded to the target object for medical image segmentation.
We have selectively sampled a few datasets that represent a wider range of modalities within radiology and non-radiology medical images.
We have worked with Kvasir-SEG~\cite{jha2020kvasir} for endoscopic images, ISIC-16~\cite{gutman2016skin} for photographic images, and CAMUS~\cite{leclerc2019deep} for radiographic images.
% If we are facilitated with more time and computing resources, we may advance to using additional datasets.

\subsection{Evaluation metrics}
We have used two evaluation metrics popular in medical image segmentation, dice score (DSC) and intersection-over-union (IoU) as:
\begin{equation}
    DSC = \frac{2*(y_{pred} \cap y_{true})}{y_{pred} + y_{true}},
\end{equation}

\begin{equation}
    IoU = \frac{y_{pred} \cap y_{true}}{y_{pred} \cup y_{true}},
\end{equation}
where $y_{pred}$ and $y_{true}$ are predicted and targeted binary masks.
For a successful adversarial attack, we compare the metrics before and after the attack.

\section{Results}

% Present your findings or results, including visual aids like graphs, tables, or screenshots, if applicable.
% \begin{table}[h]
%     % \centering
%     % \small
%     \begin{tabular}{|c|c|c|c|}
%     \hline
%         Adversarial Attack & $\epsilon$ & DSC $\uparrow$   & IoU $\uparrow$\\
%     \hline
%     Original  (W/o attack)  & - & \textbf{88.83}  & \textbf{82.72} \\
%     \hline
%         FGSM &  \begin{tabular}{c}0.01  \\  0.03  \\ 0.1 \\ 0.5  \end{tabular} &  \begin{tabular}{c} 75.08 \\  67.57  \\ 58.36 \\ 51.81  \end{tabular} & \begin{tabular}{c} 64.31 \\  55.91 \\ 45.97 \\ 39.49  \end{tabular}  \\ 
%     \hline
%         PGD &   \begin{tabular}{c}0.01 \\  0.03  \\ 0.1 \\ 0.5  \end{tabular} &  \begin{tabular}{c}71.30 \\  79.17  \\ 47.78 \\ 37.69  \end{tabular} & \begin{tabular}{c}61.84\\  71.25  \\ 37.43 \\ 27.63  \end{tabular}  \\ 
%     \hline
%     \end{tabular}
%     \caption{Comparison of Dice Score and Intersection over Union (IoU) scores across two adversarial attacks, FGSM and PGD, evaluated at four perturbation levels on ISIC-16 photographic images.}
%     \label{tab:kvasir}
% \end{table}

\begin{table*}[t]
\centering
    \begin{tabular}{|c|c|c|c|c|}
    \hline
        Dataset & Adversarial Attack & $\epsilon$ & DSC\% $\uparrow$  & IoU \% $\uparrow$  \\
    \hline \hline
     &  Original  (W/o attack)  & - & \textbf{88.83}  & \textbf{82.72} \\
    \hline
    
        Kvasir-SEG \cite{jha2020kvasir}  &  FGSM &  \begin{tabular}{c}0.01  \\  0.03  \\ 0.1 \\ 0.5  \end{tabular} &  \begin{tabular}{c} 75.08 \\  67.57  \\ 58.36 \\ 51.81  \end{tabular} & \begin{tabular}{c} 64.31 \\  55.91 \\ 45.97 \\ 39.49  \end{tabular}  \\ 
    \cline{2-5}
        &  PGD &   \begin{tabular}{c}0.01 \\  0.03  \\ 0.1 \\ 0.5  \end{tabular} &  \begin{tabular}{c}71.30 \\  79.17  \\ 47.78 \\ 37.69  \end{tabular} & \begin{tabular}{c}61.84\\  71.25  \\ 37.43 \\ 27.63  \end{tabular}  \\ 
    \hline
    
     & Original  (W/o attack)  & - & \textbf{92.27}  & \textbf{86.29} \\
    \hline
    ISIC-16 \cite{gutman2016skin} & FGSM &  \begin{tabular}{c}0.01  \\  0.03  \\ 0.1 \\ 0.5  \end{tabular} &  \begin{tabular}{c} 84.38 \\   80.23    \\ 75.84 \\ 74.05  \end{tabular} & \begin{tabular}{c} 74.79 \\ 69.02 \\ 63.14 \\ 61.51  \end{tabular}  \\ 
    \cline{2-5}
         & PGD &   \begin{tabular}{c}0.01 \\  0.03  \\ 0.1 \\ 0.5  \end{tabular} &  \begin{tabular}{c}89.34 \\  90.36  \\ 83.24 \\ 82.65 \end{tabular} & \begin{tabular}{c} 82.15\\  83.71 \\ 74.65 \\ 73.22 \end{tabular}  \\ 
    \hline
     & Original  (W/o attack)  & - & \textbf{89.87}  & \textbf{82.13} \\
    \hline
        CAMUS \cite{leclerc2019deep} & FGSM & \begin{tabular}{c}0.01 \\  0.03  \\ 0.1 \\ 0.5  \end{tabular} &  \begin{tabular}{c} 75.99 \\  73.25 \\ 71.05 \\ 48.17 \end{tabular} & \begin{tabular}{c} 63.29 \\  60.05 \\ 57.39 \\ 35.12 \end{tabular}  \\ 
    \cline{2-5}
         & PGD &   \begin{tabular}{c}0.01 \\  0.03  \\ 0.1 \\ 0.5  \end{tabular} &  \begin{tabular}{c} 46.56 \\  34.21 \\ 15.16 \\ 14.47 \end{tabular} & \begin{tabular}{c} 36.10 \\  24.85 \\ 9.22 \\ 8.6  \end{tabular}  \\ 
    \hline
    \end{tabular}
    \caption{Comparison of dice score and intersection of union scores across two adversarial attacks: FGSM and PGD  for four perturbation values in three datasets: Kvasir-SEG \cite{jha2020kvasir}, ISIC-16 \cite{gutman2016skin}, CAMUS \cite{leclerc2019deep}.  }
    \label{tab:results}
\end{table*}

\begin{figure}[h]
        % \centering
        \includegraphics[width=1\linewidth]{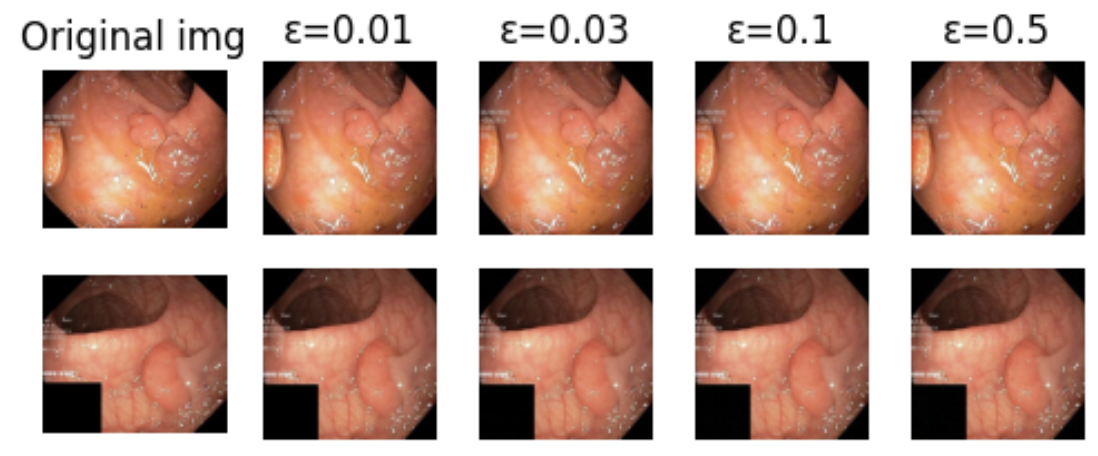}
        \caption{Comparison of original images and adversarial images generated at different perturbation levels. As the perturbation increases, the adversarial modifications become increasingly perceptible to the human eye.}
        \label{fig:fgsm-kvasir}
\end{figure}

Table \ref{tab:results} compares the performance of two different attack methods: FGSM and PGD in terms of two evaluation metrics, dice score and intersection-over-union, measured under varying perturbation levels for the given three datasets. We have chosen four perturbation levels for the study. The table shows that the attack was successful, as evidenced by the decreasing scores with increasing perturbation levels across all the datasets. 

Figure \ref{fig:fgsm-kvasir} illustrates the images generated using FGSM for the Kvasir-SEG dataset. As the perturbation size increases, the added perturbations become more noticeable, with higher visibility at $\epsilon=0.5$ compared to $\epsilon=0.01$. For additional generated adversarial images for the other two datasets, please refer to \ref{sec:generated-images}

\section{Discussion}
\paragraph{Radiological (Grayscale) images are more vulnerable.}
% Camus is an ultrasound dataset of hearts with single-channelled greyscale images with brightness values of the pixels.
Among the evaluated datasets, CAMUS experienced a more pronounced drop in DSC and IoU compared to the others under the same hyperparameter settings. A possible explanation for this discrepancy lies in the nature of the input images. While other datasets use three-channel RGB images, CAMUS consists of single-channel images representing pixel brightness.

We hypothesize that in RGB images, if an adversarial attack affects one channel of a pixel, the remaining two channels can still retain information, partially compensating for the loss — assuming those channels remain unaffected. In contrast, for an attack to fully disrupt an RGB pixel, all three channels would need to be targeted simultaneously, which is less likely.

\paragraph{Computation vs attack success vs imperceptibility.}
When comparing FGSM and PGD in terms of computation, attack success, and imperceptibility, some clear trade-offs become evident. FGSM is a single-step attack that’s computationally lightweight and easy to implement. However, its success rate tends to be lower (refer to \cref{tab:results}) than iterative methods, and it offers limited control over how noticeable the perturbations are. At higher attack strengths, it often results in visible artifacts. On the other hand, PGD builds on FGSM by applying small, incremental perturbations over multiple steps, projecting the adversarial example back within an $\epsilon$-ball, $[x-\epsilon, x + \epsilon]$, around the original input after each step. While this increases computational cost, it leads to much higher attack success rates and better imperceptibility (refer to \cref{fig:fgsm-kvasir,sec:generated-images}). Thanks to its gradual and controlled updates, PGD is widely regarded as one of the strongest and more imperceptible first-order attacks, often producing adversarial examples that are difficult to distinguish from the original images.

\section{Conclusion and Future Work}

% Summarize your key takeaways.
% Suggest potential improvements or extensions for future research.

We implemented adversarial attacks with FGSM and PGD in vision language segmentation models for medical images data. 
The findings suggest the attack on these models was successful with marginal decrease in dice score and intersection-over-union.
This is a small stepping stone towards making VLSMs robust to such data poisoning methods.

In this study, we have identified that the issues of adversarial attack also persist in the VLSM domains.
For future studies, we can research enabling the defense mechanisms against the attack. 
Also, we only have studied the white-box attacks of the models; however, we need to explore black-box attacks in which we have limited access to execution, gradients, and parameters.
We have tested it only one VLSM (i.e. CLIPSeg), but the effects of adversarial attacks must studied across other pre-existing segmentation models.
In the future, we aim to implement universal adversarial perturbation method (i.e., one adversary to attack all images in the dataset).
\bibliography{references}  
\bibliographystyle{plain}

\appendix

\section{Appendix}

\subsection{Adversarial images generated}
\label{sec:generated-images}

    \begin{figure*}[h]
        % \centering
        \includegraphics[width=\textwidth]{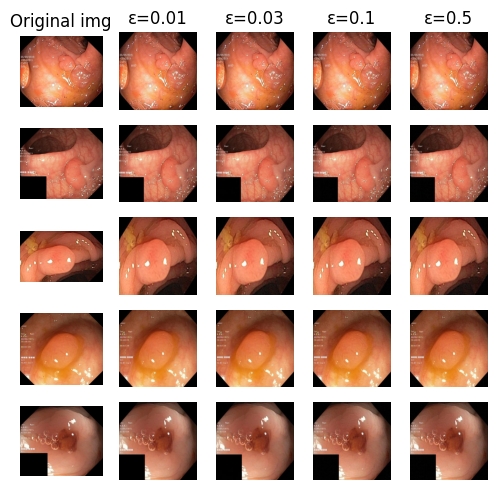}
        \caption{Comparison of original images and adversarial images generated at different perturbation levels on Kvasir dataset using PGD. As the perturbation increases, the adversarial modifications become increasingly perceptible to the human eye.}
        \label{fig:pgd-kvasir}
    \end{figure*}

    \begin{figure*}[h]
        \includegraphics[width=\textwidth]{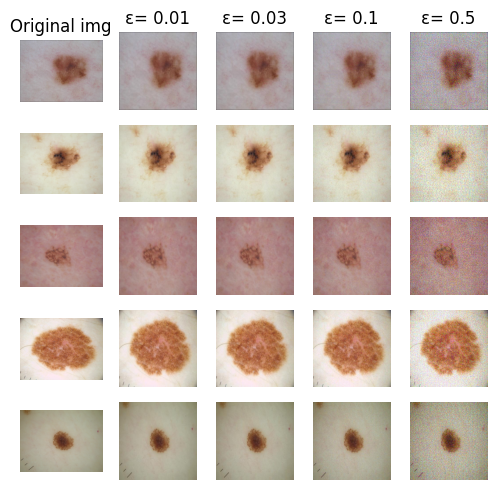}
        \caption{Comparison of original images and adversarial images generated at different perturbation levels on ISIC-16 dataset using FGSM. As the perturbation increases, the adversarial modifications become increasingly perceptible to the human eye. }
        \label{fig:fgsm-isic}
    \end{figure*}

    \begin{figure*}[h]
        \includegraphics[width=\textwidth]{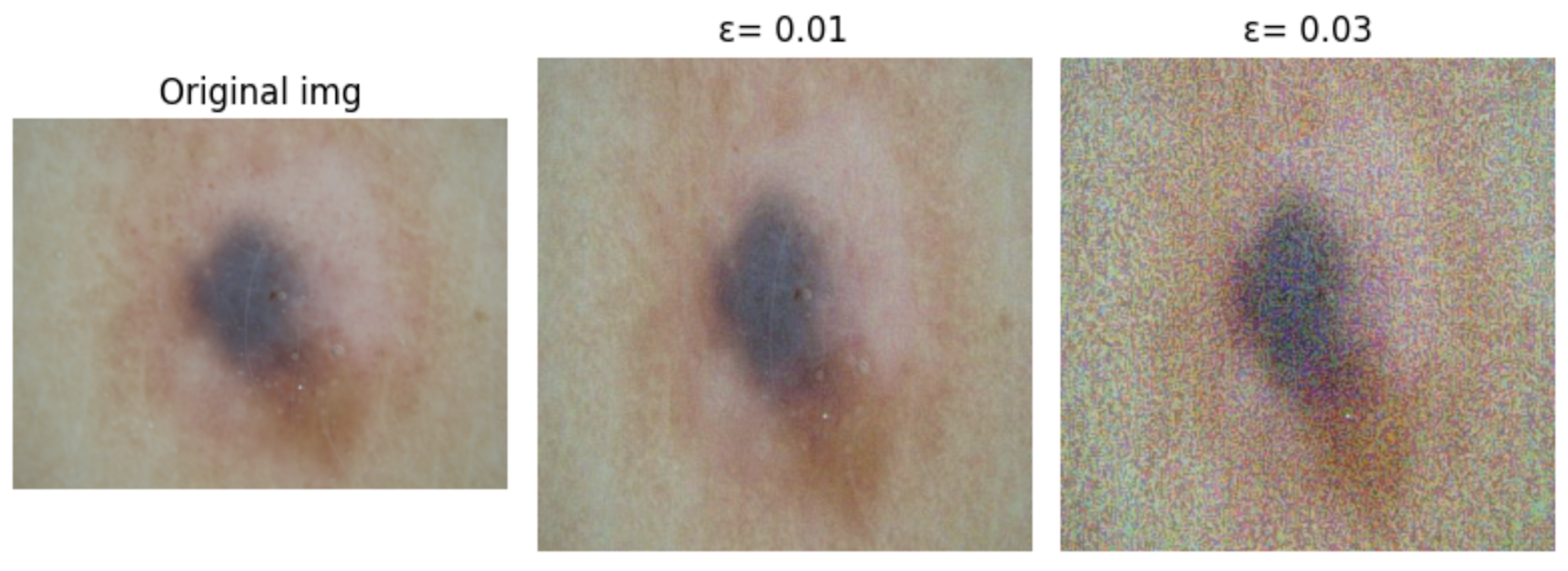}
        \caption{The provided image is a zoomed-in version highlighting the differences between the adversarial and original images. It offers a clearer comparison between the original and perturbed images generated via FGSM. This pattern is consistent across all other samples.}
        \label{fig:fgsm-isic-zoom}
    \end{figure*}

    \begin{figure*}[h]
        \includegraphics[width=\textwidth]{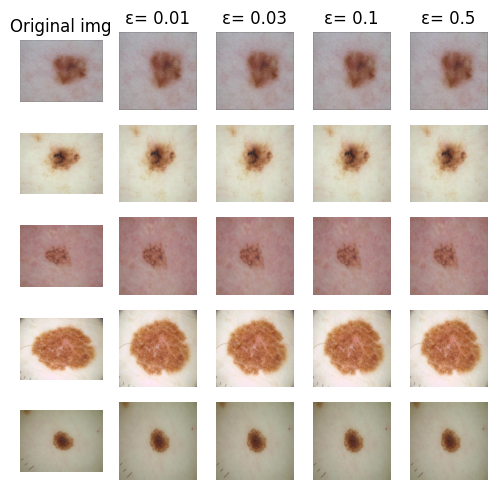}
        \caption{Comparison of original images and adversarial images generated at different perturbation levels on ISIC-16 dataset using PGD. As the perturbation increases, the adversarial modifications become increasingly perceptible to the human eye.}
        \label{fig:pgd-isic}
    \end{figure*}

    \begin{figure*}[h]
        \includegraphics[width=\textwidth]{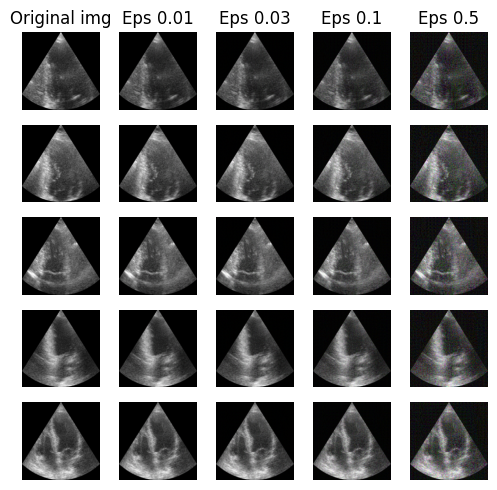}
        \caption{Comparison of original images and adversarial images generated at different perturbation levels on CAMUS dataset using FGSM. As the perturbation increases, the adversarial modifications become increasingly perceptible to the human eye.}
        \label{fig:fgsm-camus}
    \end{figure*}

     \begin{figure*}[h]
        \includegraphics[width=\textwidth]{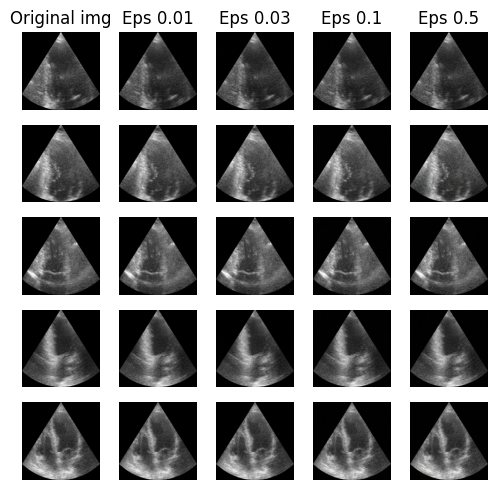}
        \caption{Comparison of original images and adversarial images generated at different perturbation levels on CAMUS dataset using PGD. As the perturbation increases, the adversarial modifications become increasingly perceptible to the human eye. }
        \label{fig:pgd-camus}
    \end{figure*}

\end{document}